\documentclass{article}

\PassOptionsToPackage{numbers, compress}{natbib}
\usepackage[final]{nips_2017}

\usepackage[utf8]{inputenc} % allow utf-8 input
\usepackage[T1]{fontenc}    % use 8-bit T1 fonts
\usepackage{hyperref}       % hyperlinks
\usepackage{url}            % simple URL typesetting
\usepackage{booktabs}       % professional-quality tables
\usepackage{amsfonts}       % blackboard math symbols
\usepackage{nicefrac}       % compact symbols for 1/2, etc.
\usepackage{microtype}      % microtypography
\usepackage{authblk}
\usepackage{graphicx}
\usepackage[titletoc,title]{appendix}

\title{Curriculum Q-Learning for Visual Vocabulary Acquisition}

\author[1,2]{\bf{{Ahmed H. Zaidi}}}
\author[1]{\bf{{Russell Moore}}}
\author[1]{\bf{{Ted Briscoe}}}
\affil[1]{Computer Laboratory, University of Cambridge}
\affil[2]{Catalyst AI Ltd., UK}
\affil[ ]{\texttt {\{ahmed.zaidi,rjm49,ted.briscoe\}@cl.cam.ac.uk}}

\begin{document}

\maketitle
\begin{abstract}
The structure of curriculum plays a vital role in our learning process, both as children and adults. Presenting material in ascending order of difficulty that also exploits prior knowledge can have a significant impact on the rate of learning. However, the notion of difficulty and prior knowledge differs from person to person. Motivated by the need for a personalised curriculum, we present a novel method of curriculum learning for vocabulary words in the form of visual prompts. We employ a reinforcement learning model grounded in pedagogical theories that emulates the actions of a tutor. We simulate three students with different levels of vocabulary knowledge in order to evaluate the how well our model adapts to the environment. The results of the simulation reveal that through interaction, the model is able to identify the areas of weakness, as well as push students to the edge of their ZPD. We hypothesise that these methods can also be effective in training agents to learn language representations in a simulated environment where it has previously been shown that order of words and prior knowledge play an important role in the efficacy of language learning. 
\end{abstract}

\section{Introduction}
With the rise of machine learning and the tasks such as automated teaching and assessment, there is an increased interest in understanding how machine learning models can be grounded in theories of language acquisition. Additionally, with the abundance of learner data in archive and generation, we now have an avenue through which we can not only evaluate our theories of learning, but also explore whether these theories can be used to train agents for the purpose of general AI. 

Language Acquisition is a multidisciplinary field that includes linguistics, psychology, neuroscience, philosophy, and more recently computer science. At the intersection of language acquisition and pedagogy lie theories of educational practices for language learners, including for example, an optimal curriculum for both L1 and L2 learners. A {\it{curriculum}} is a guide that helps teachers decide what content to present and the order of which it needs to be presented. The aim of a curriculum is to provide a highly structured method of introducing concepts in order to maximise the rate of learning. 

The idea of a curriculum to facilitate the rate of learning has been discussed from the perspective of {\it{animal training}} \cite{skinner1958teaching,peterson2004day}, where it is defined as {\it{shaping}}. It has also been referenced in an educational framework \cite{bruner1960process} where the author introduces the idea of a {\it{spiral curriculum}}, a process by which complex information is first presented in a simplified manner and then revisited at a more difficult level later on. Similarly Vygotsky, from the view of language acquisition, introduces the idea of {\it{scaffolding}} in order to provide contextual support for more complex ideas using simplified language or visuals. All of these concepts have been discussed in different fields but reference the same underlying idea of presenting information in a structured manner in order to exploit prior knowledge. 

Bruner \cite{bruner1961act} argues that the role of the teacher is not to present information by rote learning but rather facilitate the learning process in order to teach students to become {\it{active learners}}: put simply, they are ``learning to learn''. There are many factors that teachers need to consider when constructing a curriculum to achieve this goal, namely the {\it{difficulty}} and {\it{appropriateness}} of content. 

Difficulty is measured relative to the {\it{zone of proximal development (ZDP)}}, introduced by Vygotsky, which is a representation of what a learner is capable of achieving without help, with {\it{some}} help, and of concepts that are beyond the learner's current ability. Appropriateness is a measure of whether content being presented is within the ZPD or, in the case of scaffolding, comprises material from within the ZPD. 

Determining difficulty and appropriateness is traditionally a very laborious and resource intensive task which entails experts conducting focus groups and analysis to decide where a particular question or topic sits in the curriculum. This method is not only inefficient, it also assumes a static curriculum for all students. 

To address these limitations, we propose the use of reinforcement learning (RL) in order to learn an optimal policy and curriculum for each student in the task of visual vocabulary acquisition. Through this, we also discuss the similarities between the properties and features of RL and those of language acquisition. We evaluate our models by simulating three types of student at different levels of proficiency {\it{(beginner, intermediate, and advanced)}}. We find that the system is able to identify the difference in proficiency and adapt its curriculum to reflect.

Previous uses of RL in pedagogy include \cite{beck2000advisor} where it is used to teach students arithmetic, aiming to minimise the time taken to answer questions. \cite{iglesias2009learning,iglesias2003experience} teach students database design using Q-learning. Both \cite{beck2000advisor} and \cite{iglesias2009learning,iglesias2003experience} evaluate results on simulated students. \cite{martin2004agentx} use RL for maths while \cite{tetreault2006comparing} use it for physics. However, as far as we know, no previous work has been done in the space of visual lexical acquisition where the principles of RL have explicitly been related to theories of language acquisition.    

The importance of curriculum learning in training deep learning models and agents has also been discussed by \cite{bengio2009curriculum} where its use is shown to facilitate the generalisation as well as the rate of convergence and training of deep learning networks. \cite{hermann2017grounded} also illustrate the need for some form of curriculum to improve the rate of learning for agents in a 3D simulation. However, it is worth noting that no explicit RL is used to model curriculum by either \cite{bengio2009curriculum} or \cite{hermann2017grounded}.

%Bruner (1960) - spiral curriculum 
%Bruner (1961) - discovery learning
%Wood et al., 1976) - scaffolding block reconstruction

\section{Curriculum Q-Learning}

In order to automate the process of curriculum learning for visual vocabulary acquisition, we must first identify the key components of our RL system. The agent in this task is the automated tutor that must learn what information to present to the student. The environment is the student who is interacting with the agent.  

We assume that the student interacting with the tutor is a learner of English who has reached a given level on the \emph{Common European Framework of Reference} (CEFR) scale. CEFR is an international standard for describing language ability, using a six point scale, from A1 for beginners, up to C2 for those who have mastered a language. 

The RL algorithm used by our proposed system is Q-Learning, an {\it{off-policy}} algorithm for Temporal Difference (TD) Learning. Q-Learning can be defined as follows:

\begin{equation}
Q(s,a) \leftarrow Q(s,a) + \alpha [r + \gamma \max_{a'}Q^\pi(s',a') - Q(s,a)] 
\end{equation}

where $Q(s,a)$ is the Q-value of a state $s$ and action $a$ tuple. The $\alpha$ is the learning rate and $\gamma$ is the discount factor. $\gamma$ models the fact that future rewards are less valuable than immediate rewards at a given time $t$. 

A policy $\pi$ maps states $s$ to actions $a$. The aim of the Q-Learning algorithm is to find an optimal policy $\pi$ such that it maximises the long-term cumulative reward. The policy achieves this by acting greedily and taking the action that presents the maximum Q-value given the state such that $\max_{a\in A}Q^\pi(s_{t},a)$.  

In action selection, there is a trade-off between {\it{exploiting}} what you have learnt so far and {\it{exploring}} other state-action tuples. In this task we model that using $\epsilon$-greedy. This means the policy will, for most part, select the actions that provide the highest estimated future reward given the state. However, with a probability of $1-\epsilon$, an action will be selected randomly and independently from a uniform distribution. Action selection is usually drawn from a Q-Table which is a table that stores all state-action Q-values.

In this task, a policy can be viewed as a curriculum as it decides what should be shown and in what order. In order to learn a curriculum for vocabulary acquisition, we incorporate two models, the {\it{CEFR level model}} and {\it{word level model}}. The CEFR level model has 6 states which are defined by the 6 CEFR levels. The actions are whether the student should progress to the next level, stay in the current level, or go back a level. The word level model has two states: active (show the word), inactive (hide the word). The actions are remain in the current state and toggle state. This architecture ensures that there is also an estimated long-term reward associated with showing a student a particular word.  

Modelling reward is often viewed as a challenging task in RL. For this application, a student is rewarded negatively (-1) for getting a question correct and positively (+1) for getting it incorrect. The motivation behind using these values is grounded in how we learn. As the RL model acts greedily and takes the action with the maximum reward, if we review a concept we understand, then we are not gaining in terms of knowledge by reviewing it again and thus its value should be reduced. Alternatively, if we get a question wrong, the benefit of reviewing that word is higher and thus we should increase the associated Q-value.  

To evaluate the students' understanding, we present a word in a form of an image. The objective for the students is to describe the image, and based on their response, the Q-Learning algorithm and thus the policy is updated. A valid response is defined by the target word associated with the image or a synonym of that target word which is automatically generated by looking at the top 10 nearest words to the target word in a pre-trained {\it{word2vec}} model \cite{mikolov2013efficient}. The use of images was motivated by it's inexpensive nature of producing evaluation material. Additionally, there are countless studies that indicate the effectiveness of images for learning \cite{verdi1997organized}.

\section{Experiments}

%\subsection{Visual Vocabulary}
For the CEFR level model, we use a learning rate $\alpha$ of $0.1$, a discount rate $\gamma$ of $0.9$ and an $\epsilon$ value of $0.95$. The word level model uses an $\alpha$ of $0.1$, a $\gamma$ of $0.9$ and an $\epsilon$ value of 1 in order to prevent words randomly going into an inactive state. 

%\subsection{Simulated Students}
%Reinforcement learning (RL) systems frequently suffer from the `cold start' problem: it takes a long time to accumulate enough data to begin optimising a policy in a meaningful way.  How many episodes are needed depends on the problem and algorithm being used, but tens or hundreds of thousands of passes is fairly typical.  It is thus impractical to cold-start an RL system that is aimed at interacting meaningfully with human subjects: we might expect the system to take months or years to converge on a useful policy using observations returned at human speed, and in the meantime the human guinea-pigs would be subjected to all manner of (seemingly to them) random exploratory behaviour.

%In this context, then, some kind of warm-up phase is required.  In this work, \emph{simulated humans} are used as the guinea-pigs.  Such `simulants' are agent-based models designed to work in a human-like way.

% A simple model of vocabulary acquisition is used.  Using annotated corpora of speech and writing from L2 students of English, we extract frequency counts of relevant vocabulary items at each CEFR level.  It is assumed that a user who is correctly using a vocabulary word in their speech or writing, would also recognise an image of that item if it were shown to them. %TODO actually about twice this many...

%\subsubsection{Probability model}

To evaluate the performance of our system, we simulated three types of students at varying levels of proficiency: {\it{beginner, intermediate and advanced}}. In this case, we modelled the student's probability of getting a question correct as a negated Gompertz distribution:

\begin{equation}
	P(success~|~u,q) = 1 - \exp(-b~\exp(-c(l(q) - l(u))))
\end{equation}

where $l(u)$ denotes the level of user $u$ calibrated to a scale of $\left[0,6\right]$. Each integer in the scale represents a corresponding CEFR level from A1 to C2 (\emph{e.g.} 0 $\rightarrow$ A1, 1 $\rightarrow$ A2, \emph{etc.}).  $l(q)$ represents the level of an item $q$ (i.e. a word which must be guessed from an image) calibrated to the same scale.  The parameter $b$ determines the probability of success when student and item level match. This is set to $ln(0.75)$ to model a `typical' pass rate of 75\%. The calibrated curve is shown in Appendix \ref{gompertz}. The curve is flatter at the lower end as students may be expected to be comfortable with most of the material at lower CEFR levels than their own, whereas at higher levels, their ability is more uncertain. We ran simulations where each student had 100 interactions with the system. An interaction can be defined as when a student responds to a question. 

\begin{figure}[!tbp]
  \centering
  \begin{minipage}[b]{0.49\textwidth}
    \includegraphics[width=\textwidth]{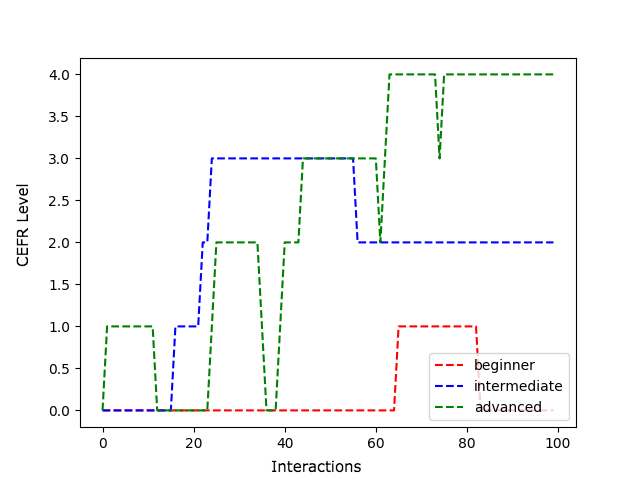}
    \caption{CEFR levels determined by the agent for students of varying levels of proficiency over 100 interactions}
    \label{fig:level}
  \end{minipage}
  \hfill
  \begin{minipage}[b]{0.49\textwidth}
    \includegraphics[width=\textwidth]{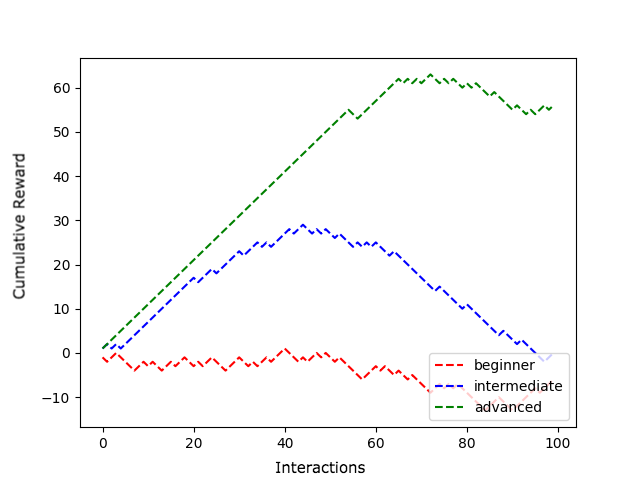}
    \caption{Cumulative reward earned by students of varying levels of proficiency from the agent over 100 interactions}
    \label{fig:reward}
  \end{minipage}
\end{figure}

\subsection{Results}

The results from Figure \ref{fig:level} show how the agent responds to the various proficiency levels. The beginner student remains relatively constant around A1 and A2 which is reflective of the student's current level. The intermediate student continually increases in CEFR level until level 3 (B2). The advanced student, although is tested with material beneath the actual level of proficiency, eventually reaches an advanced or higher CEFR level. We can also see that the agent tutor pushes the student to what can be interpreted as the edge of their ZDP.  Figure \ref{fig:reward} illustrates how the cumulative reward of the students varies for students at different proficiencies. The curve experiences a downward slope as the students reach their current level of vocabulary and are now being pushed to understand material beyond their scope. 

\section{Discussion}

We have shown through the use of simulations, that we can effectively model a personalised curriculum for vocabulary acquisition  using Q-Learning. Figure \ref{fig:level} and Figure \ref{fig:reward} show clear indications of varying agent behaviour for students at different levels of lexical proficiency. However, beyond that, we have set up a framework that can be used in the future to extrapolate the difficulty and appropriateness of new material. The system will serve as a test bed that will yield metrics to determine where the content fits in the curriculum. Although this is foundational work, it lays the building blocks for future pedagogically inspired RL architectures. 

Through this work, we have also shown that there are many similarities between the principles of RL and theories of language acquisition. Specifically, parallels can be drawn between the concept of $\epsilon$-greedy and Krashen's Input Hypothesis or the {\it{i+1}}. The Input Hypothesis states that students progress their learning by comprehending language that is slightly above their current language level. The interactions between the agent and the environment in RL is analogous to the social interaction approach to language acquisition, specifically equal importance of input and output. We also use the Q-Learning algorithm as opposed to the SARSA algorithm mainly due to the properties of Q-Learning that ensure an "optimal path" is followed i.e. the minimum number of steps to reach our goal (language fluency). 

However, there is scope for substantial extensions in this space. Deploying the system on-line in order to collect user data will allow us to validate and improve our existing models. Incorporating memory and spaced repetition learning in order to optimise the policy and emulate cognitive processes is also an important extension that may have a great impact on the learning output. Using deep learning models to approximate the Q-value will allow the system to capture additional signals pertinent to language acquisition. Additionally, moving towards an adaptive reward model that reflects difficulty to encourage memory retention. 

All of these models can also be applied to agents instead of students. As discussed previously, \cite{hermann2017grounded} indicated the need for a curriculum in order to effectively train an agent in the simulated environment. Creating a dynamic environment guided by a curriculum grounded in pedagogically inspired RL may result in improved learning rates for the agent.   

\subsubsection*{Acknowledgements}

We thank Wenchao Chen who helped develop the back-end of our web-based platform. 
\newpage
\bibliographystyle{unsrt}

\begin{thebibliography}{10}

\bibitem{skinner1958teaching}
Burrhus~Frederic Skinner.
\newblock Teaching machines.
\newblock {\em Science}, 128(3330):969--977, 1958.

\bibitem{peterson2004day}
Gail~B Peterson.
\newblock A day of great illumination: Bf skinner's discovery of shaping.
\newblock {\em Journal of the Experimental Analysis of Behavior},
  82(3):317--328, 2004.

\bibitem{bruner1960process}
Jerome~S Bruner.
\newblock {\em The process of education:[a searching discussion of school
  education opening new paths to learning and teaching]}.
\newblock Vintage Books, 1960.

\bibitem{bruner1961act}
Jerome~S Bruner.
\newblock The act of discovery.
\newblock {\em Harvard educational review}, 1961.

\bibitem{beck2000advisor}
Joseph Beck, Beverly~Park Woolf, and Carole~R Beal.
\newblock Advisor: A machine learning architecture for intelligent tutor
  construction.
\newblock {\em AAAI/IAAI}, 2000:552--557, 2000.

\bibitem{iglesias2009learning}
Ana Iglesias, Paloma Mart{\'\i}nez, Ricardo Aler, and Fernando Fern{\'a}ndez.
\newblock Learning teaching strategies in an adaptive and intelligent
  educational system through reinforcement learning.
\newblock {\em Applied Intelligence}, 31(1):89--106, 2009.

\bibitem{iglesias2003experience}
Ana Iglesias, Paloma Martinez, and Fernando Fern{\'a}ndez.
\newblock An experience applying reinforcement learning in a web-based adaptive
  and intelligent educational system.
\newblock 2003.

\bibitem{martin2004agentx}
Kimberly~N Martin and Ivon Arroyo.
\newblock Agentx: Using reinforcement learning to improve the effectiveness of
  intelligent tutoring systems.
\newblock In {\em Intelligent Tutoring Systems}, pages 564--572. Springer,
  2004.

\bibitem{tetreault2006comparing}
Joel~R Tetreault and Diane~J Litman.
\newblock Comparing the utility of state features in spoken dialogue using
  reinforcement learning.
\newblock In {\em Proceedings of the main conference on Human Language
  Technology Conference of the North American Chapter of the Association of
  Computational Linguistics}, pages 272--279. Association for Computational
  Linguistics, 2006.

\bibitem{bengio2009curriculum}
Yoshua Bengio, J{\'e}r{\^o}me Louradour, Ronan Collobert, and Jason Weston.
\newblock Curriculum learning.
\newblock In {\em Proceedings of the 26th annual international conference on
  machine learning}, pages 41--48. ACM, 2009.

\bibitem{hermann2017grounded}
Karl~Moritz Hermann, Felix Hill, Simon Green, Fumin Wang, Ryan Faulkner, Hubert
  Soyer, David Szepesvari, Wojtek Czarnecki, Max Jaderberg, Denis Teplyashin,
  et~al.
\newblock Grounded language learning in a simulated 3d world.
\newblock {\em arXiv preprint arXiv:1706.06551}, 2017.

\bibitem{mikolov2013efficient}
Tomas Mikolov, Kai Chen, Greg Corrado, and Jeffrey Dean.
\newblock Efficient estimation of word representations in vector space.
\newblock {\em arXiv preprint arXiv:1301.3781}, 2013.

\bibitem{verdi1997organized}
Michael~P Verdi, Janet~T Johnson, William~A Stock, Raymond~W Kulhavy, and Polly
  Whitman-Ahern.
\newblock Organized spatial displays and texts: Effects of presentation order
  and display type on learning outcomes.
\newblock {\em The Journal of Experimental Education}, 65(4):303--317, 1997.

\end{thebibliography}

\newpage
\begin{appendices}

\section{Curriculum Q-Learning System Overview}

\begin{figure}[h]
\begin{center}
\includegraphics[width=13cm]{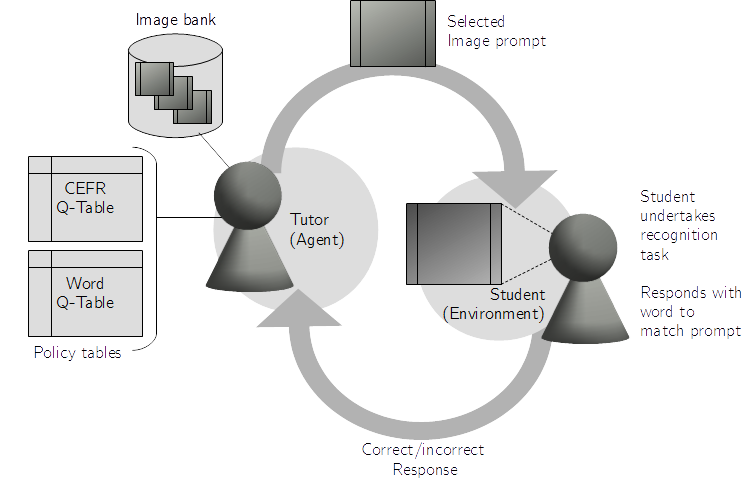}
% \fbox{\rule[-.5cm]{0cm}{4cm} \rule[-.5cm]{4cm}{0cm}}
\end{center}
\caption{Overview of the system.  A simulated student takes the place of a human actor in our study.}
\label{fig:sys_overview}
\end{figure}

\section{Negated Gompertz Curve}
\label{gompertz}

\begin{figure}[h]
\begin{center}
\includegraphics[width=10cm]{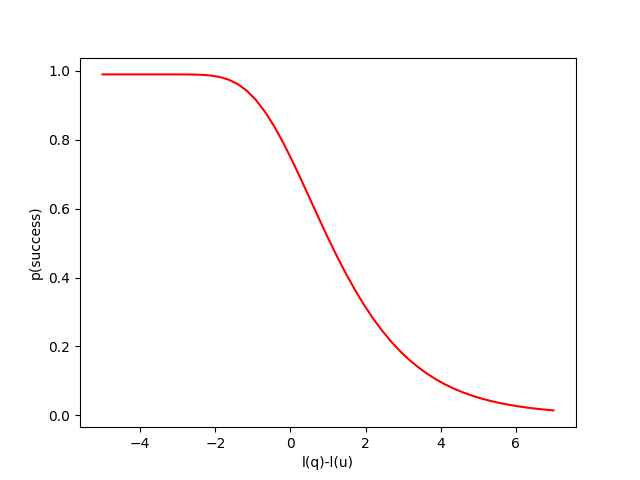}
% \fbox{\rule[-.5cm]{0cm}{4cm} \rule[-.5cm]{4cm}{0cm}}
\end{center}
\caption{Gompertz curve used as a model to simulate student success probabilities.}
\label{fig:prob}
\end{figure}
\newpage
\section{Preview of Web-based Curriculum Q-Learning}

\begin{figure}[h]
\begin{center}
\includegraphics[width=14cm]{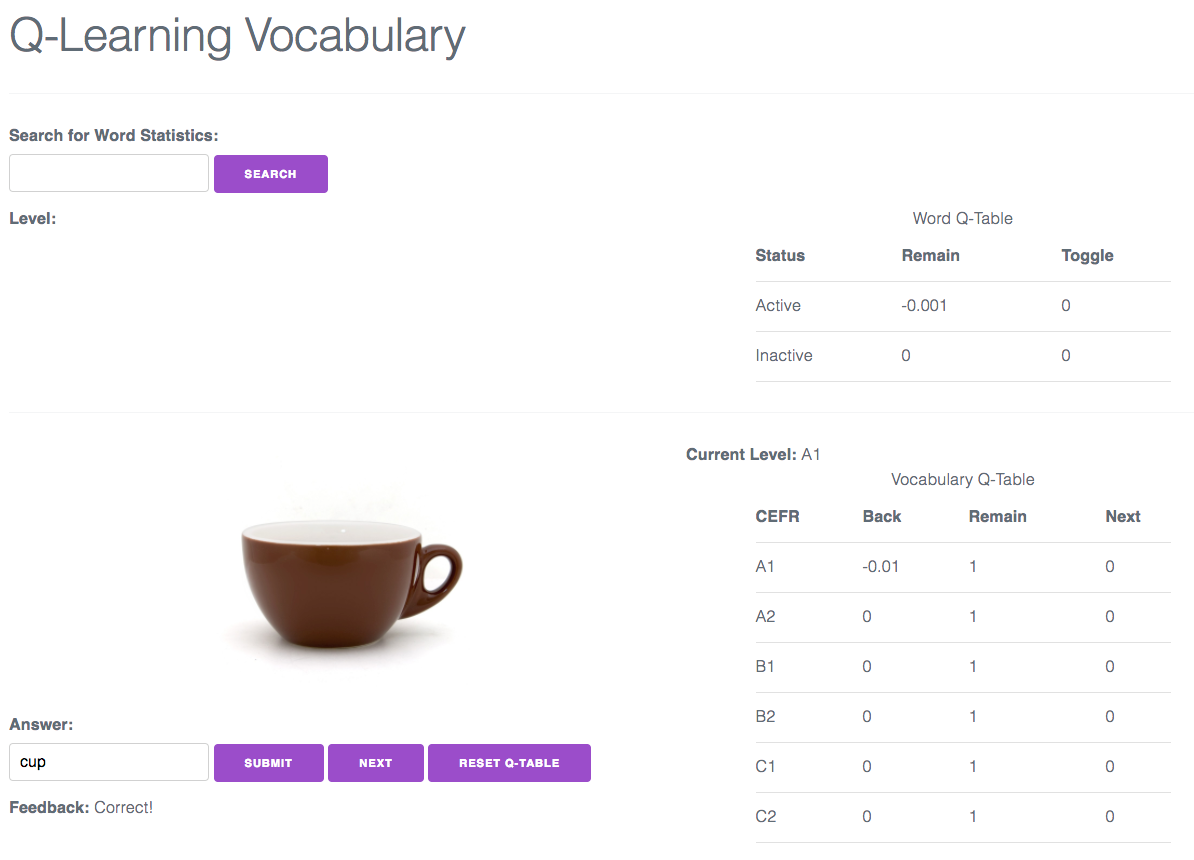}
% \fbox{\rule[-.5cm]{0cm}{4cm} \rule[-.5cm]{4cm}{0cm}}
\end{center}
\caption{A preview of the web-based Curriculum Q-Learning platform.}
\label{fig:prob}
\end{figure}

\end{appendices}
\end{document}